\documentclass[10pt,conference]{IEEEtran}
\IEEEoverridecommandlockouts

\usepackage{amsmath}
\usepackage{amssymb}
\usepackage{amsthm}
\usepackage{rotating}
\usepackage{adjustbox}
\usepackage{cite}
\usepackage{hyperref}
\usepackage{orcidlink}

\newcommand{\C}{\mathbb{C}}
\newcommand{\R}{\mathbb{R}}
\newcommand{\K}{\mathbb{K}}
\newcommand{\Real}{\mathcal{R}}
\newcommand{\Imag}{\mathcal{I}}
\newcommand{\real}{\mathcal{R}}

\DeclareMathOperator{\argmax}{argmax}
\DeclareMathOperator{\sign}{sign}
\DeclareMathOperator{\st}{s.t.}

\title{Perturbing the Phase: Analyzing Adversarial Robustness of Complex-Valued Neural Networks}
\author{\IEEEauthorblockN{Florian Eilers \orcidlink{0000-0003-0726-3287}}
\IEEEauthorblockA{\textit{Department of Computer Science} \\
\textit{University of Münster}\\
Münster, Germany}
\and
\IEEEauthorblockN{Christof Duhme \orcidlink{0009-0004-1853-2862}}
\IEEEauthorblockA{\textit{Department of Computer Science} \\
\textit{University of Münster}\\
Münster, Germany}
\and
\IEEEauthorblockN{Xiaoyi Jiang \orcidlink{0000-0001-7678-9528}}
\IEEEauthorblockA{\textit{Department of Computer Science} \\
\textit{University of Münster}\\
Münster, Germany}}

\begin{document}

\maketitle

\begin{abstract}
Complex-valued neural networks (CVNNs) are rising in popularity for all kinds of applications. To safely use CVNNs in practice, analyzing their robustness against outliers is crucial. One well known technique to understand the behavior of deep neural networks is to investigate their behavior under adversarial attacks, which can be seen as worst case minimal perturbations. We design Phase Attacks, a kind of attack specifically targeting the phase information of complex-valued inputs. Additionally, we derive complex-valued versions of commonly used adversarial attacks. We show that in some scenarios CVNNs are more robust than RVNNs and that both are very susceptible to phase changes with the Phase Attacks decreasing the model performance more, than equally strong regular attacks, which can attack both phase and magnitude.
\end{abstract}

\textbf{\textcopyright 2026 IEEE}. Personal use of this material is permitted.  Permission from IEEE must be obtained for all other uses, in any current or future media, including reprinting/republishing this material for advertising or promotional purposes, creating new collective works, for resale or redistribution to servers or lists, or reuse of any copyrighted component of this work in other works.

\section{Introduction}

While most deep learning architectures are still real-valued neural networks (RVNNs), complex-valued neural networks (CVNNs) have lately risen in popularity. 
Their natural ability to deal with complex-valued inputs make them feasible for a number of applications that deal with complex-valued images including safety critical applications such as MR imaging \cite{cole2021analysis, vasudeva2022compressed}, autonomous driving \cite{moon2025mounting} and many more \cite{Fuchs2025, Nguyen2023, zhang2024complex}. 

Deep neural networks have been shown to be vulnerable to drastic output changes upon small input perturbations. The robustness of models against small distribution changes can be investigated by adversarial attacks. They can be understood as a "worst case" minimal perturbation that maximizes the change of the output.

While many works have investigated how RVNNs react to adversarial attacks, no such investigation has been made for CVNNs. Besides introducing how to calculate adversarial attacks in the complex domain, we present a comparative study analyzing the robustness of RVNNs and CVNNs. To this end, we present a novel attack - the Phase Attack, that specifically targets the phase of a complex-valued input. The phase information is often neglected in visualization and is hard to understand, thus perturbations to it are especially prone to be unnoticed.
To further evaluate the impact of phase perturbation, we also design its counterpart, the Magnitude Attack, which only targets the magnitude but leaves the phase unchanged. 
We show, that Phase Attacks are the most effective attacks, when compared to regular attacks and Magnitude attacks, showing the susceptibility of both CVNNs and RVNNs to phase perturbation, when dealing with complex valued inputs.

To summarize, we contribute:
\begin{itemize}
    \item A new kind of attack: Phase Attacks, that are specifically suited to attack complex-valued inputs in real-world applications and analysis of its effectiveness.
    \item An comparative analysis on the adversarial robustness between CVNNs and RVNNs.
    \item A toolbox in PyTorch to apply all presented approaches to the complex domain.
    The code can be found at \url{https://zivgitlab.uni-muenster.de/ag-pria/cv-attacks-phase-attacks}.
\end{itemize}

\section{Related Work}

\subsection{Adversarial Attacks}
Since the definition of "adversarial attacks" \cite{szegedy2013intriguing} the effect of minor perturbations fooling deep neural networks has been analyzed extensively \cite{goodfellow2014explaining, madry2018towards, wong2020fast,dong2018boosting, lin2020nesterov, wang2021enhancing, xie2019improving}. More recently, deep neural networks were utilized to craft more powerful adversarial attacks against other models \cite{liang2022adversarial}. Additionally, defense algorithms have been proposed as counter measures against adversarial attacks \cite{tramer2017ensemble, papernot2016distillation, cohen2019certified}, however many of them can be circumvented \cite{athalye2018obfuscated}. Adversarial training describes the counter measure, where adversarial attacks are applied during training and has been shown very effective \cite{madry2018towards, wong2020fast, shafahi2019adversarial}. Besides these practical works, many works have analyzed the robustness of deep neural networks and the effects, that lead to adversarial vulnerability \cite{stutz2019disentangling, yin2019fourier}.

\subsection{Complex-Valued Neural Networks}
Since the early days of deep learning, CVNNs have been a point of interest \cite{aizenberg1971about, hirose2003complex, nitta1997extension, yang1994complex}. More recently, deep CVNNs have been introduced with basic building blocks \cite{trabelsi2018deep}, as well as advances in model architectures \cite{yang2020complex, eilers2023building, lee2022complex}, theoretical understanding \cite{tan2022real, chen2023spectral} and hardware optimization \cite{zhang2021optical, bai2025tops}. They have lately risen in popularity for a multitude of applications \cite{zhong2023real,chen2023spectral,liu2023pixelwise,xing2023phase,yakupouglu2024comparison}, including safety critical applications, such as MRI image processing \cite{virtue2017better, cole2021analysis, dou2025mri} or autonomous driving \cite{moon2025mounting}. The robustness to noise of analog CVNNs has been investigated in the context of optical neural networks \cite{ron2023noise}. 

Zhou et al. \cite{zhou2023phase} have shown, that the phase information of the Fourier transform of adversarial attacks crafted for real-valued images is pivotal. They however do not use this knowledge to craft an attack that focuses on the phase and do only investigate the effect of the phase of the Fourier transform of natural images processed by RVNNs, rather than also incorporating complex-valued images, which offer a phase information in image space.

In a theoretical work, Neaccsu et al. \cite{neaccsu2022design} present a Lipschitz bound for robustness against adversarial attacks of CVNNs and derive a shallow model architecture as well as a training routine to satisfy this Lipschitz bound, however their approach is based on nonnegative neural networks, which are unsuitable for deep neural networks.
They do not compare the robustness of RVNNs and CVNNs.

Lastly, the closest related work to our work is a work by Yeats et al. \cite{yeats2021improving}. They investigate how gradient regularization has different effects on adversarial robustness of CVNNs and RVNNs and conclude, that gradient regularization is more effective in CVNNs than in RVNNs. However, they only investigate this phenomenon on real-valued image datasets. We are interested in the setting, where the input data is complex-valued, since that is a primary use case for CVNNs. 

To the best of our knowledge, no study has so far compared the adversarial robustness of RVNNs and CVNNs for complex-valued inputs or investigated the impact of phase perturbations to complex-valued inputs.

\section{Gradient-Based Adversarial Attacks}\label{sec:rv_aa}

An adversarial attack can loosely be defined as an imperceivable perturbation to input data of a deep learning model that significantly alters the output of the model.
While adversarial attacks are mostly studied for images, the general framework is applicable to any input vector.
We thus use the terms input (vector), which has entries, instead of speaking of images and pixels specifically.

Adversarial attacks can formally be formulated as a restricted optimization problem.
Let $f$ be some trained (real-valued) deep learning model, $X\in\R^n$ its input with label $Y$ and some objective function $l$.
Then the adversarial input $Z^*$ is defined as:
\begin{align}\label{eq:opt_grad_real}
    Z^*=\argmax_{Z\in \R^n} l(f(Z), Y) \ \ \st ||Z-X||_\infty < \epsilon
\end{align}
The bound $\epsilon>0$ is a small real number that is tuned depending on the dataset at hand. A greater value makes the attack stronger, but also leads to a more perceivable perturbation.

To solve this restricted optimization problem efficiently, gradient based optimization algorithms can be utilized. The Fast Gradient Sign Method (FGSM) \cite{goodfellow2014explaining} makes use of a single gradient step. The attacked input $Z_\text{FGSM}$ is then defined as:
\begin{align}
	Z_\text{FGSM} = X + \epsilon \ \text{sign}\left(\nabla l(f(X))\right)
\end{align}
where sign describes the point wise sign of the gradient vector.
This basic single step approach can now be extended by, instead of starting with $X$, starting at a random position in the search space $B_{\epsilon,X}= \{Z \in \R^n| \ ||Z-X||_\infty < \epsilon\}$. This approach is called Faster FGSM  (FFGSM) \cite{wong2020fast}, the attacked input $Z_\text{FFGSM}$ is then defined as:
\begin{align}
    Z_\text{FFGSM} &= \mathcal{P}_{B_{\epsilon,X}} \left( X + U + \epsilon \ \text{sign}\left(\nabla l(f(X + U))\right)\right)
\end{align}
where $U\in\R^n$ denotes a random vector from the uniform distribution $\mathcal{U}(B_\epsilon(0))$ and $\mathcal{P}_{B_{\epsilon,X}}$ denotes the projection onto $B_{\epsilon,X}$, ensuring the $\epsilon$-bound.

This can be further extended by not just going a single gradient step, but iteratively refining the attacked input.
This was first proposed by Kurakin et al. \cite{kurakin2018adversarial} (Basic Iterative Method) and was later extended to the Projected Gradient Descent by including a random starting position \cite{madry2018towards}.
To ensure consistency in our notation, we call it Iterative FGSM (IFGSM).
For some $0 < \alpha < \epsilon$ (often $\alpha = \frac{\epsilon}{4}$) and some maximum step $m$, the attacked input is defined as $Z_\text{IFGSM} = Z_m$ and then:
\begin{align}
    Z_0 &= X + U\\
    Z_{t+1} &= \mathcal{P}_{B_{\epsilon,Z_t}} \left( Z_t + \alpha \ \text{sign}\left(\nabla l(f(Z_t))\right)\right)
\end{align}
Lastly, a momentum term can be added to the gradient calculation. This leads to Momentum IFGSM (MIFGSM) \cite{dong2018boosting} and is (with a momentum controlling parameter $0\leq \beta \leq 1$) again defined as $Z_\text{MIFGSM} = Z_m$ with:
\begin{align}
    Z_0 &= X + U; \ M_0 = 0 \\
    M_{t} &= M_{t-1} + \beta \ \text{sign}\left(\nabla l(f(Z_t))\right) \\
    Z_{t+1} &= \mathcal{P}_{B_{\epsilon,Z_t}} \left( Z_t + \alpha \ \text{sign}\left(\nabla l(f(Z_t)) + M_{t}\right)\right)
\end{align}
Many variants and improvements to these four base methods have been proposed, such as using Nesterov Momentum \cite{lin2020nesterov}, Variance Tuning \cite{wang2021enhancing}, Input Diversity \cite{xie2019improving} and many more \cite{akhtar2018threat}. Since these are all specific improvements to the base methods above, we focus on the four base types to compare robustness between RVNNs and CVNNs.

\section{Complex-Valued Gradient-Based Attacks}\label{subsec:grad_cvaa}

In this section, we describe how to design complex-valued adversarial attacks.
For the adaption we can rely on Wirtinger calculus \cite{wirtinger1927formalen}, a well known tool for optimization of real-valued functions (such as loss functions) with complex-valued inputs. With $\Real(z), \Imag(z)$ being the real- and imaginary parts of $z\in\C$ and $i$ the complex unit, the Wirtinger derivatives for a function $f: \C \to \K$ with $\K=\C \text{ or } \R$ are defined as:
\begin{align}
    \frac{\delta f}{\delta z} &:= \frac{1}{2} \left(\frac{\delta f}{\delta \Real(z)} - i \frac{\delta f}{\delta \Imag(z)}\right) \\ 
    \frac{\delta f}{\delta \overline{z}} &:= \frac{1}{2} \left(\frac{\delta f}{\delta \Real(z)} + i \frac{\delta f}{\delta \Imag(z)}\right)\label{eq:wirtinger_zconj}
\end{align}
The Wirtinger derivative w.r.t. the complex conjugate $\frac{\delta f}{\delta \overline{z}}$ points in the direction of the strongest ascent and the magnitude equals the amount of change in that direction \cite{hirose2003complex}, just like the gradient in the real domain. Thus we can use it to solve the optimization problem posed for adversarial attacks in the complex domain.
Let $X\in \C^n$ be a complex-valued input with label $Y$ to some (real-valued or complex-valued) model $f$ with objective function $l$ and $\epsilon > 0$ the perturbation bound. The complex-valued version of \eqref{eq:opt_grad_real} then states:
\begin{align}\label{eq:opt_grad}
	&Z^*=\argmax_{Z\in \C^n} l(f(Z), Y) \\
	&\st ||Z-X||_\infty < \epsilon \nonumber
\end{align}
where the complex infinity norm is defined as:
\begin{align}
    ||z||_{\infty} & := \max_{1\leq j \leq n}(|z_j|)
\end{align}
We now briefly discuss how to adapt all presented methods from Section~\ref{sec:rv_aa} using Wirtinger derivatives. Since MIFGSM is the most advanced version of the previously introduced methods, we will define $\C$MIFGSM in detail, all other versions can then be achieved by simplification. 

To this end, let $U_\C$ be a random input of size of $X$ drawn from the complex uniform distribution on the $\epsilon$-ball around $0$: $\mathcal{U}_\C(B_{\epsilon,0})$.
Additionally, let $0 \leq \beta \leq 1$ and $0 < \alpha < \epsilon$.
Then, for some max step size $m$, $\C$MIFGSM is calculated as $Z^* = Z_m$ by:
\begin{align} \label{eq:cmifgsm_init}
    Z_0 &= X + U_\C; \ M_0 = 0 \\
    M_{t} &= M_{t-1} + \beta \ \text{sign}\left(\nabla_{\overline{z}} l(f(Z_t))\right) \\
    Z_{t+1} &= \mathcal{P}_{B_{\epsilon,Z_t}} \left( Z_t + \alpha \ \text{sign}\left(\nabla_{\overline{z}} l(f(Z_t)) + M_{t}\right)\right) \label{eq:cmifgsm_end}
\end{align}
With $\phi_Z$ the phase of $Z\in\C$ its sign is defined as:
\begin{align}
	\sign(Z)=\sign(|Z|\exp(i\phi_Z)) = \exp(i\phi_Z)
\end{align}
Additionally, $\nabla_{\overline{z}}$ denotes the Wirtinger gradient with partial derivatives defined as in \eqref{eq:wirtinger_zconj} and $\mathcal{P}_{B_{\epsilon,Z_t}}$ denotes the projection onto $B_{\epsilon,Z_t})$.
From \eqref{eq:cmifgsm_init}-\eqref{eq:cmifgsm_end} we can define $\C$IFGSM, by setting $\beta=0$, $\C$FFGSM by also setting $\alpha=\epsilon$ and $m=1$ and finally $\C$FGSM by setting $U_\C = 0$.

\section{Phase Attacks}\label{phase_cvaa}
The attacks presented in Section~\ref{subsec:grad_cvaa} are straightforward extension from the real domain, however they do not consider the unique nature of the complex domain.
When handling complex data in practical applications, it is common practice to only visualize the magnitude information, since the phase information is normally very noisy and hard to understand.
Even if the phase information is visualized, perturbations within the phase are hard to detect due to the heavy noise, especially in regions of low signal magnitude.
The basic idea of our proposed Phase Attacks is to specifically target the phase of the complex-valued input and leave the magnitude unperturbed to specifically analyze, how sensitive a deep neural network is to phase perturbations.

\begin{figure}[t]
    \begin{minipage}{0.33\linewidth}
        \includegraphics[width=\linewidth]{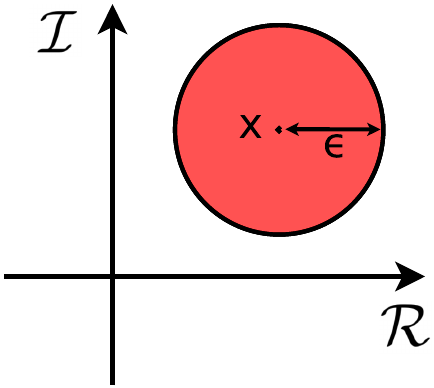}
    \end{minipage}%
    \begin{minipage}{0.67\linewidth}
    \begin{center}
    \hfill
    \includegraphics[width=.49\linewidth]{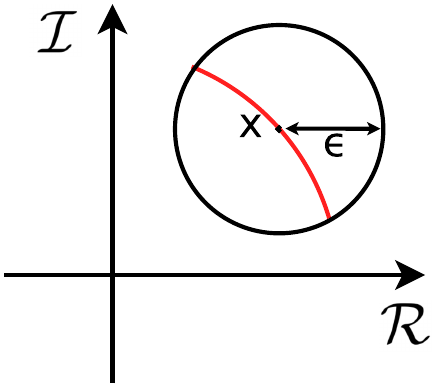}
    \hfill
    \includegraphics[width=.49\linewidth]{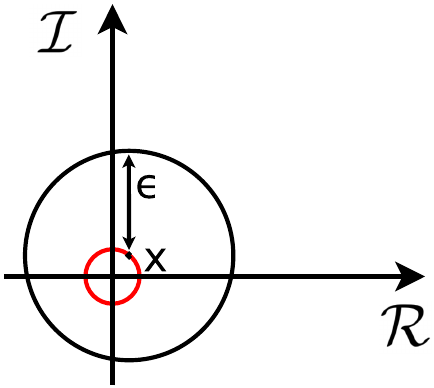} \\

    \hfill
    \includegraphics[width=.49\linewidth]{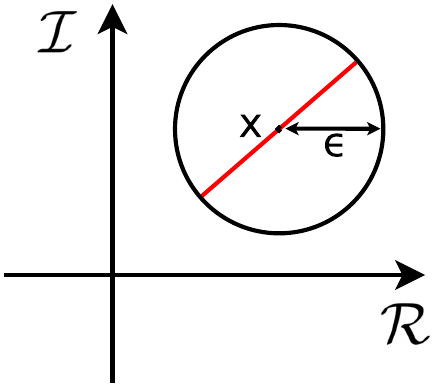}
    \hfill
    \includegraphics[width=.49\linewidth]{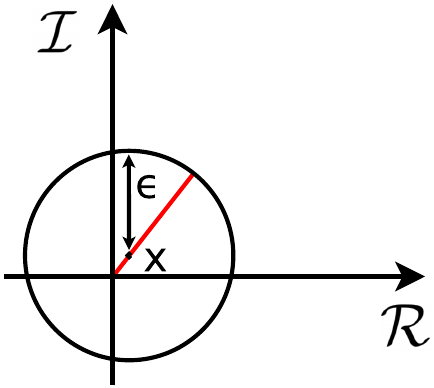}
    \end{center}
    \end{minipage}
    \caption{Visualization of search spaces for adversarial attack optimization problems.
    Black circle shows epsilon ball around value $X$, search space in red.
    Left: Classical adversarial attack \eqref{eq:opt_full}, center top/bottom: Phase Attack/Magnitude Attack for $X$ far from zero, right top/bottom: Phase Attack/Magnitude Attack for $X$ close to zero (\eqref{eq:opt_angle}/\eqref{eq:opt_prob_mag}).}
    \label{fig:search_space}
\end{figure}

To define the Phase Attack we keep the $\epsilon$-restraint from Section~\ref{subsec:grad_cvaa} but additionally add the constraint, that we do not perturb the magnitude of the input.
Let $X \in \C^n$ be an input to be perturbed with label $Y$, $f$ a (real- or complex-valued) model to predict $Y$ from $X$ and $l$ a suitable loss function.
We then have to solve:
\begin{align}\label{eq:opt_full}
    &Z^*=\argmax_{Z\in \C^n} l(f(Z), Y) \\
    &\st ||Z-X||_\infty < \epsilon \text{ and } |Z| \underset{p.w.}{=} |X| \nonumber
\end{align}
This is equivalent to solving the following:
\begin{align}
    & Z^* = |X| \exp(i(\phi_X + \phi_Z^*)) \text{  with }  \label{eq:opt_angle} \\
    & \phi_Z^* = \argmax_{\phi_Z \in [-\pi, \pi)} l(f( |X| \cdot \exp(i (\phi_X + \phi_Z))), Y) \nonumber \\
    & \st \ 2|X| \underset{p.w.}{\leq} \epsilon \text{ or } |\phi_X-\phi_Z| \underset{p.w.}{<} 2 \arcsin\left(\frac{\epsilon}{2|X|}\right) \nonumber 
\end{align}
\begin{proof}
We have to show that for every $x \in \C$
\begin{align}
    & \bigg\{z\in \C \bigg| |z| = |x|  \text{ and }|z-x| <\epsilon \bigg\} \nonumber\\
    = &\bigg\{z \in \C \bigg| |z|=|x| \text{ and }  \nonumber \\
    & \bigg( 2|z| \leq \epsilon \text{ or } |\phi_x - \phi_z| <  2 \arcsin\left(\frac{\epsilon}{2|x|}\right) \bigg) \bigg\}
\end{align}
Let $x \in \C$, we now have to differentiate between two cases:\\
\underline{1. case}: $2|x|\leq \epsilon$:\\
$|z| = |x|$ implies $2|z|\leq \epsilon$, which leads to the left set being included in the right set. Additionally, we have $|x-z|<|x|+|z|=2|x|<\epsilon$ and thus the opposite inclusion.

\noindent \underline{2. case}: $2|x| > \epsilon$:\\
We have to show, that (if $|x|=|z|$):
\begin{align}
    |x-z|<\epsilon \Leftrightarrow |\phi_x - \phi_z| < 2 \arcsin\left(\frac{\epsilon}{2|x|}  \right)
\end{align}
For $x, z$ satisfying the left inequality, define $\phi_y := \phi_x - \phi_z$ and thus $x=|z|\exp(i(\phi_z + \phi_y))$. Then it holds, that
\begin{align}
& \ |x-z|<\epsilon \nonumber\\
\Leftrightarrow & \ ||z|\exp(i\phi_z + \phi_y) - |z|\exp(i\phi_z) |  < \epsilon \\
\Leftrightarrow & \ |(\exp(i\phi_y) - 1) \ z|  < \epsilon\\
\Leftrightarrow & \ |(\exp(i\phi_y) - 1)| \ |z|  < \epsilon\\
\Leftrightarrow & \ |(\exp(i\phi_y) - 1)|  < \frac{\epsilon}{|z|}
\end{align}
We can now use, that $\exp(i\phi_y)$ and $1\in \C$ form an isosceles triangle with angle $\phi_y$ and leg length $1$. Then $|(\exp(i\phi_y) - 1)|$ is the length of the base of this triangle. Thus
\begin{align}
    |(\exp(i\phi_y) - 1)| = 2\left|\sin\left(\frac{\phi_y}{2}\right)\right|
\end{align}
Since the $|\sin|$ is an even function, applying $\arcsin$ to both sides of the following inequality yields
\begin{align}
    & 2\left|\sin\left(\frac{\phi_y}{2}\right)\right| < \frac{\epsilon}{|z|} \nonumber \\
    \Leftrightarrow & \ |\phi_y| < 2 \arcsin\left(\frac{\epsilon}{2|z|}\right)
\end{align}
Thus, combining all equivalences yields 
\begin{align}
& \ |x-z|<\epsilon \nonumber\\
    \Leftrightarrow & \ |\phi_x - \phi_z| < 2 \arcsin\left(\frac{\epsilon}{2|z|}\right)
\end{align}
This concludes the proof. 
\end{proof}
Note that we apply both restrictions point wise to the input and the logical 'or' also needs to be understood pointwise.
In other words: For any entry, one of the two pointwise restrictions needs to hold.
The first restriction means that we can freely perturb the phase of entries with magnitude close enough to $0$, while the second restriction shows that for entries with increasing magnitude the allowed phase perturbations decrease.
We have visualized this in Fig.~\ref{fig:search_space}.

\subsection{Optimization Algorithms for Phase Attacks}

We present an optimization algorithm to solve the optimization problem in \eqref{eq:opt_angle}. Our approach is inspired by the existing methods presented in Section~\ref{subsec:grad_cvaa}.
Just like $\C$FGSM, the general idea of our optimization algorithm is to determine an optimal direction of change for every entry (which is guided by the gradient) and then make a maximum step into that direction.

We describe the one step attack in detail, which can then be extended to a multi step attack similarly to how $\C$FGSM can be extended to $\C$MIFGSM by starting in a random position in the search space, applying the steps multiple times and replacing the gradient by a momentum enhanced gradient. We call the respective concrete implementations of Phase Attacks PFGSM, PFFGSM, PIFGSM and PMIFGSM.

Since the formulation in \eqref{eq:opt_angle} consists of two restrictions, we have to differentiate which restriction is limiting the perturbation of a specific entry.
As highlighted before, if $2|X|\leq \epsilon$ holds in an entry, we can freely perturb the phase, if this does not hold, we have to restrict the phase perturbation to $2\arcsin(\epsilon / (2|X|))$.
In the first case, we differentiate further between two scenarios: The Wirtinger derivative of that entry points into or out of the search space (which is just a sphere in this case).
All three scenarios are highlighted below and visualized in Fig.~\ref{fig:optim_algo}.

\subsubsection{Optimization for Unrestricted Phase Perturbations} 
\begin{figure}[t]
    \begin{center}    
    \includegraphics[width=.25\linewidth]{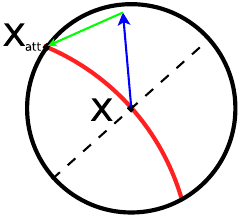} \hspace{.04\linewidth} \vline \hspace{.04\linewidth} 
    \includegraphics[width=.25\linewidth]{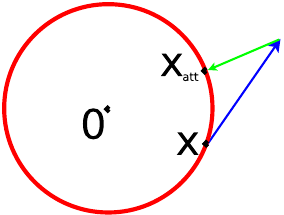}\hspace{.04\linewidth} \vline \hspace{.04\linewidth} 
    \includegraphics[width=.25\linewidth]{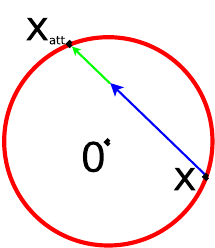}
    \caption{Visualization of optimization steps for Phase Attack optimization problems. Search space in red (see Fig.~\ref{fig:search_space}), Wirtinger gradient in blue. Left: restricted phase perturbation \eqref{eq:optim_step_restricted}, center: freely perturbed phase with outward pointing gradient \eqref{eq:optim_step_free_out}, right: freely perturbed phase with inward pointing gradient \eqref{eq:inward}.}
    \label{fig:optim_algo}
    \end{center}
\end{figure}

\ \\
We assume, we can freely perturb the phase (i.e., the condition $2|X| \leq \epsilon$ holds for an entry).
Thus, we want to perturb the phase of an entry of an input to model $f$.
Let $X\in \C$ be the corresponding value of that entry and $\frac{\delta f}{\delta \overline{x}} \in \C$ its Wirtinger derivative w.r.t. the complex conjugate.
We now have to differentiate between two scenarios: When looking at the circle with magnitude $|X|$, then $\frac{\delta f}{\delta \overline{x}}$ either points outwards that circle from $X$ or inwards (see Fig.~\ref{fig:optim_algo}). 
Precisely, we have to distinguish between:
\begin{align} \label{eq:out}
	\text{Pointing outwards: } &\Real\left(\frac{\delta f}{\delta \overline{x}} \overline{X}\right) \geq 0 \\
	\text{Pointing inwards: } &\Real\left(\frac{\delta f}{\delta \overline{x}} \overline{X}\right) < 0 \label{eq:in}
\end{align}
Note that $\Real(a\overline{b}) = \langle (\Real(a), \Imag(a)), (\Real(b), \Imag(b)) \rangle_{\R^2}$ holds for $a,b \in \C$, which leads to an easily understandable interpretation of the inequality in \eqref{eq:out}. The formulation above is analogous to a formulation we need later for the optimization in the restricted case. 

\noindent \textbf{Optimization in case of outward pointing derivative} \\
\noindent In this case, we use the directional information (i.e., the sign) and the strength information (i.e., the magnitude) of the gradient to generate an attack. A regular gradient descent step would lead to $X + \frac{\delta f}{\delta \overline{x}}$, however $|X + \frac{\delta f}{\delta \overline{x}}| > |X|$, thus we need a projection onto our search space, the circle of magnitude $|X|$. To summarize, the Phase Attack in the unrestricted case with an outward-pointing gradient can be calculated as:
\begin{align}\label{eq:optim_step_free_out}
    X_{\text{att}} = \sign\left(X + \frac{\delta f}{\delta \overline{x}}\right) |X|
\end{align}

\noindent \textbf{Optimization in case of inward pointing derivative} \\
\noindent For the inward pointing derivative we use the directional information of the derivative and then use the maximum possible perturbation in that direction. If $\phi_{\delta \overline{x}}$ denotes the phase of $\frac{\delta f}{\delta \overline{x}}$, we calculate the following value:
\begin{align}\label{eq:inward}
    &X_{\text{att}} = X + \alpha \exp(i\phi_{\delta \overline{x}}) \\
    &\text{with } \alpha > 0 \text{ s.t. } |X_{\text{att}}| = |X| \nonumber
\end{align}
This formulation has a closed form solution for $\alpha$ with:
\begin{align}
	\alpha = -2|X|\cos(\phi_{\delta \overline{x}} - \phi_X)
\end{align}
fulfilling $\alpha > 0$ for $|\phi_{\delta \overline{x}} - \phi_X| \in [\frac{\pi}{2}, \frac{3\pi}{2}]$, which is equivalent to the gradient pointing inwards.
\begin{proof}
We have to find $\alpha > 0$ such that 
\begin{align}
    \big| |X| \exp(i\phi_X) + \alpha \exp(i\phi_{\delta \overline{z}})\big| = |X|
\end{align}
Simple arithmetic in the complex domain yields
\newcommand{\ba}{\phi_{\delta \overline{z}} - \beta}
\begin{align}
    & \big| |X| \exp(i\phi_X) + \alpha \exp(i\phi_{\delta \overline{z}})\big| = |X| \\
    \Leftrightarrow & \big| |X| + \alpha \exp(i(\phi_{\delta \overline{z}} - \beta))\big| = |X| \\
    \Leftrightarrow & \sqrt{(|X| + \alpha \cos(i(\phi_{\delta \overline{z}} - \beta)))^2 + \alpha^2 \sin^2(\ba)} \nonumber \\ & = |X| \\
    \Leftrightarrow & |X|^2 + 2\alpha \cos(\ba) |X| + \alpha^2 \cos^2(\ba) \nonumber \\
    & + \alpha^2 \sin^2(\ba)  = |X|^2
\end{align}
With the use of the Pythagorean identity $\cos^2+\sin^2=1$ for the equivalence between Equations \ref{cvaa_eq:proof_pyth1} and \ref{cvaa_eq:proof_pyth2} and by solving a quadratic polynomial (in $\alpha$) in the last equivalence we get
\begin{align}
    & |X|^2 + 2\alpha \cos(\ba) |X| + \alpha^2 \cos^2(\ba) \nonumber \\
    & + \alpha^2 \sin^2(\ba)  = |X|^2 \label{cvaa_eq:proof_pyth1} \\
    \Leftrightarrow & 2\alpha \cos(\ba) |X| + \alpha^2 = 0  \label{cvaa_eq:proof_pyth2}\\
    \Leftrightarrow & \alpha = -2 |X|\cos(\ba)
\end{align}
It holds, that $\alpha > 0$ iff $\cos(\ba) < 0$, which is equivalent to the gradient pointing inwards as required in this case. This concludes the proof.
\end{proof}

\subsubsection{Optimization for Restricted Phase Perturbations} \ \\
In the case where we can only perturb the phase in a restricted way, we determine the perturbation direction (increasing or decreasing the phase) and then perturb the phase maximally in that direction.
Note that according to \eqref{eq:opt_angle} the maximum perturbation of the phase is $2 \arcsin\left(\frac{\epsilon}{2|X|}\right)$ (for those entries, that are restricted).
We can determine the direction as follows: 
\begin{align}\label{eq:dec_phase1}
    & \text{We have to increase the phase, if }\Imag\left(\frac{\delta f}{\delta \overline{x}} \overline{X}\right) > 0
\end{align}
and vice versa. Thus, we get the perturbed phase by:
\begin{align}\label{eq:optim_step_restricted}
    X_{\text{att}} = |X| \exp\left(i \phi_X +/- 2 \arcsin\left(\frac{\epsilon}{2|X|}\right)\right)
\end{align}
where we use an addition or subtraction based on \eqref{eq:dec_phase1}.

\subsection{Magnitude Attack}

To be able to analyze the impact of the Phase Attack properly, we design the contrary adversarial attack as well: 
The Magnitude Attack, e.g. an adversarial attack, that allows perturbation of the magnitude, while the phase stays constant.

To this end, let $X \in \C^n$ be an input to be perturbed with label $Y$, $f$ a (real- or complex-valued) model to predict $Y$ from $X$ and $l$ a suitable loss function. For the Magnitude Attack, we then have to solve:
\begin{align}
    &Z^*=\argmax_{Z\in \C^n} l(f(Z), Y) \\
    &\st ||Z-X||_\infty < \epsilon \text{ and } \phi_Z \underset{p.w.}{=} \phi_X \nonumber
\end{align}
This is equivalent to solving the following:
\begin{align}\label{eq:opt_prob_mag}
    & Z^* = \tilde{Z}^* \exp(i \phi_X) \text{ with } \\
    & \tilde{Z}^* = \argmax_{\tilde{Z} \in \R_{\ge 0}^n} l(f(\tilde{Z} \exp(i \phi_X)), Y) \nonumber \\
    & \st || \ \tilde{Z} - |X| \ ||_\infty < \epsilon \nonumber
\end{align}
$\tilde{Z} \in \R_{\ge 0}^n$ is the magnitude of the desired output, restricting it to be greater than or equal to zero guarantees that the phase cannot flip, i.e. that the phase is not altered by $\pi$. This is visualized in Fig.~\ref{fig:search_space}.

To optimize the Magnitude Attack, we once again determine the direction of an optimization step and then update the magnitude in that direction with the maximum allowed perturbation.
For an optimization step of step size $\epsilon$ this leads to:

\begin{align}\label{eq:optim_mag}
    X_{\text{att}} = \max\left(0, X + \epsilon \sign\left( \real\left(\frac{\delta f}{\delta \overline{x}} \overline{X}\right)\right)\right) e^{i\phi_X}
\end{align}
Note, that the same logic applies as in \eqref{eq:out} and \eqref{eq:in} to determine the direction of optimization step.

\section{Experimental Results}
We conduct experiments on two image classification datasets from two different domains which often use CVNNs: the  S1SLC\_CVDL PolSAR \cite{asiyabi2023complex} and FastMRI Prostate \cite{tibrewala2024fastmri} (preprocessed as proposed by \cite{rempe2024tumor}) datasets and train a real-valued and a complex-valued ResNet \cite{he2016deep} and ConvNeXt \cite{liu2022convnet} model on each dataset, where we used initialization from RVNNs for the CVNNs \cite{eilers2025initializing}. We used their approach EQ without additional pretraining of the RVNNs. 
Both models were trained with a learning rate of $10^{-4}$ for $50$ epochs using the Adam optimizer \cite{kingma2015adam} on a cross entropy loss function with batch sizes of $64$ and $200$ on the Prostate and PolSAR datasets, respectively.
Early stopping was used, if the validation F1 score did not improve for 10 epochs, we stop the training and use the model with the best validation F1 score for the evaluation on the test set.
We trained all models with and without additional adversarial training \cite{madry2018towards} which is known to improve adversarial robustness.

With these experiments, we aim to compare adversarial robustness between RVNNs and CVNNs.
As highlighted above, we are specifically interested in how perturbations of the phase information in complex-valued inputs influence the model performance.
Since real-valued datasets do not possess phase information, this is a phenomenon limited to the complex domain.
To get insight into those two aspects --- the general adversarial robustness and specifically the robustness to Phase Attacks --- we conduct two separate sets of experiments on all trained models in Section~\ref{subsec:realvscv} and Section~\ref{subsec:basevsphase}.
All results shown are normalized to the model performance without adversarial attacks, since we are not interested in comparing model performance, but rather robustness.

\subsection{Comparing Robustness of Real- and Complex-Valued Models} \label{subsec:realvscv}

\begin{figure*}[th!]
    \begin{center}    
    \begin{turn}{90} \hspace{7mm} PolSAR dataset \end{turn}
    \includegraphics[width=.9\linewidth]{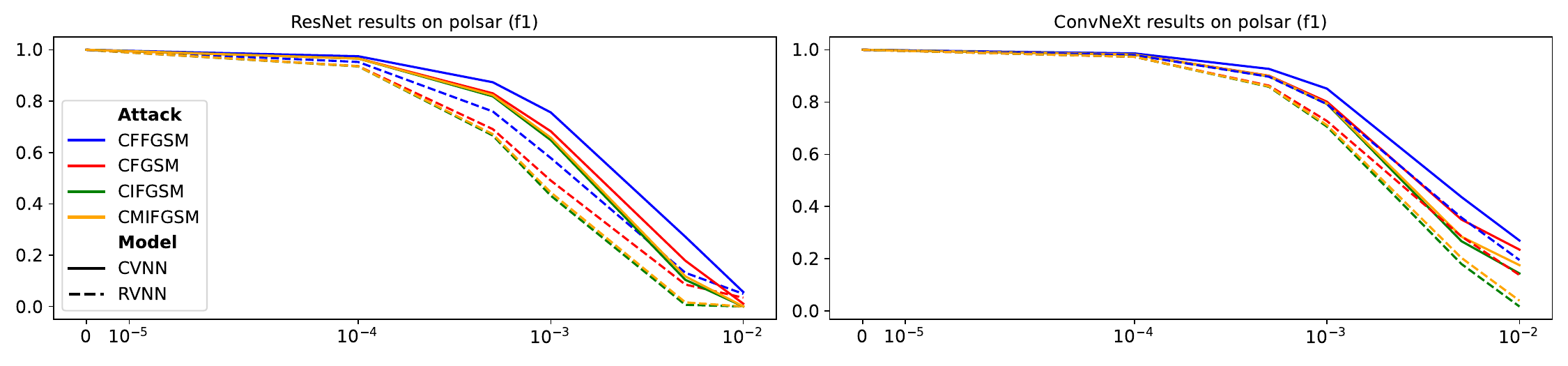}
    
    \begin{turn}{90} \hspace{7mm} Prostate dataset \end{turn}
    \includegraphics[width=.9\linewidth]{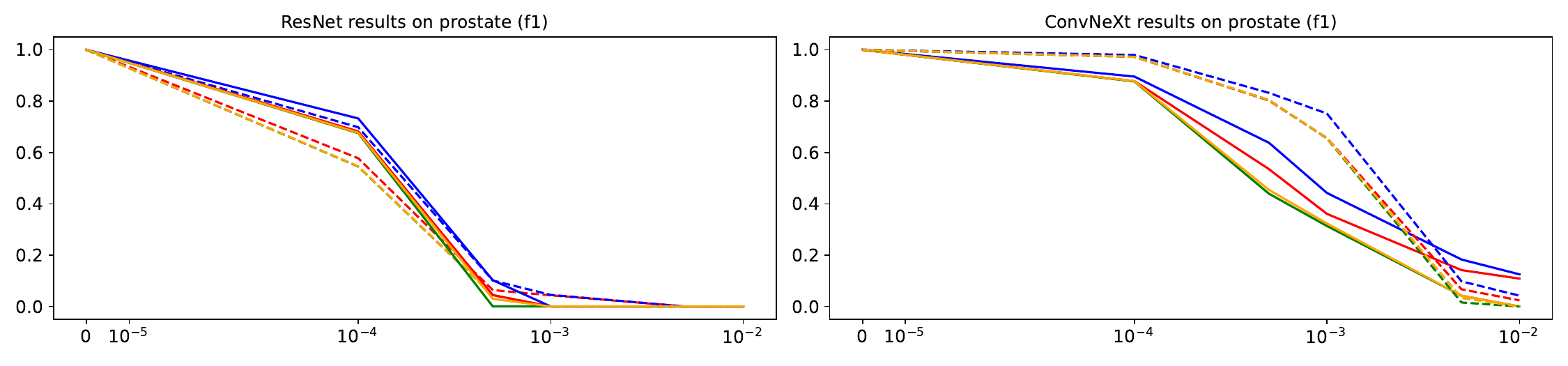}
        
    \begin{turn}{90} \hspace{7mm} PolSAR dataset \vspace{10mm} \end{turn}
    \includegraphics[width=.9\linewidth]{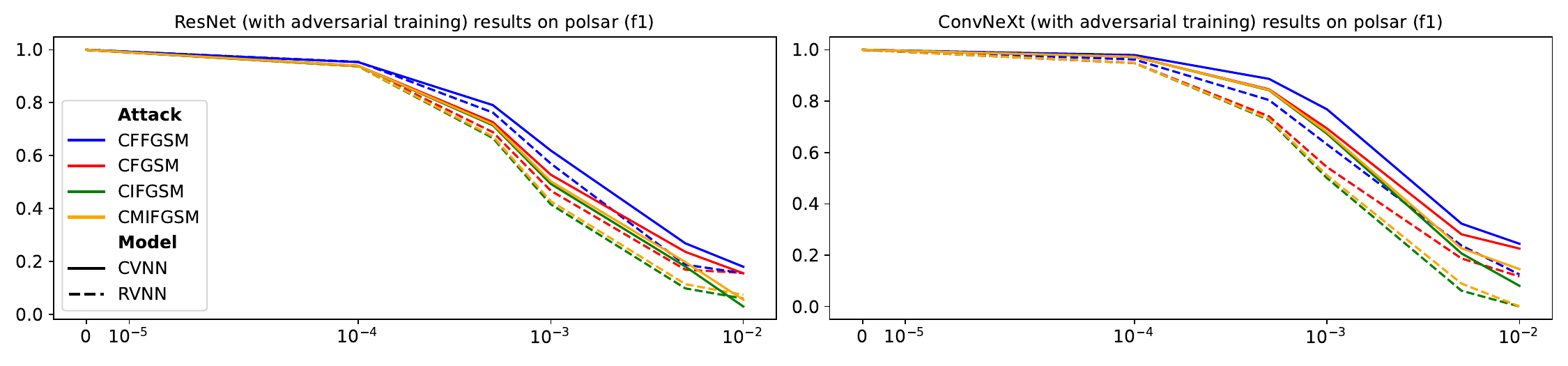}
    
    \begin{turn}{90} \hspace{7mm} Prostate dataset \vspace{10mm} \end{turn}
    \includegraphics[width=.9\linewidth]{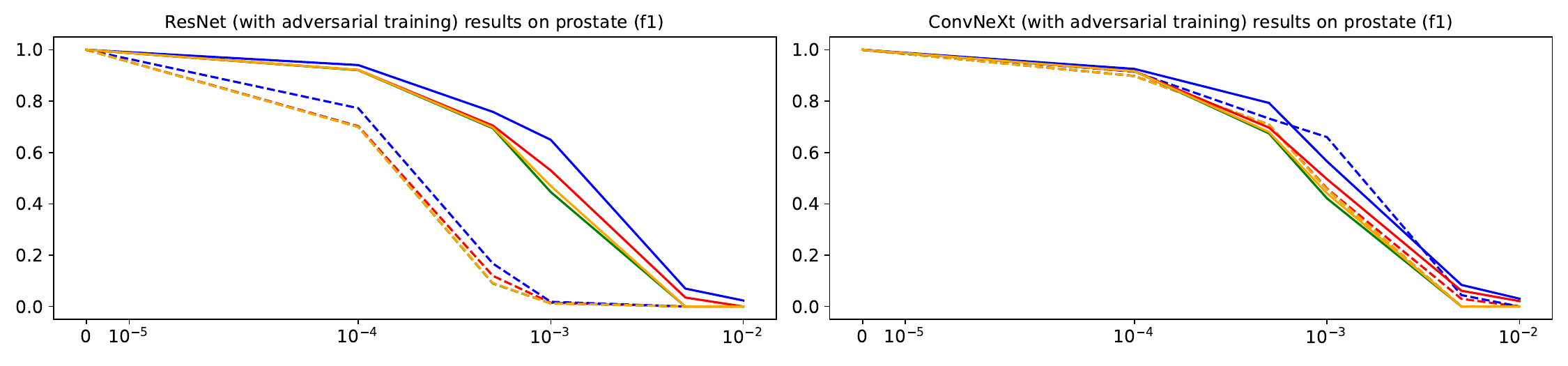}
    
    \caption{Comparison of robustness against (unrestricted) adversarial attacks of RVNNs and CVNNs on the FastMRI Prostate and PolSAR datasets without (Top two rows) and with (Bottom two rows) additional adversarial training.
    Left: ResNet, Right ConvNeXt.
    All plots show (normalized) performance against $\epsilon$-restriction.
    Linestyle: straight line: CVNNs, dashed line: RVNNs.
    Colors differentiate attack algorithms as shown in legend.}
    \label{fig:cvvsreal}
    \end{center}
\end{figure*}
In this experiment we compare the robustness of RVNNs and CVNNs when presented with the straightforward adaptations of adversarial attacks as presented in Section~\ref{subsec:grad_cvaa}.

On the PolSAR dataset (Fig.~\ref{fig:cvvsreal}, top row) it shows that the complex-valued ResNet and ConvNeXt are both more robust to adversarial attacks. While both RVNNs and CVNNs start to decrease when exposed to an attack restricted by $\epsilon=0.005$ both RVNNs decrease faster. The CVNNs are more robust at all noise levels.

On the Prostate dataset (Fig.~\ref{fig:cvvsreal}, second row) the results for the ResNet model are similar, all methods lead to a significant reduction in model performance with a sharp drop after $\epsilon=0.0001$ and the CVNNs show a higher robustness to the adversarial attacks than the RVNNs. The ConvNeXt model shows the opposite behavior, for this model, the RVNNs are much more robust to the adversarial attacks at almost all noise levels.

\begin{figure*}[th!]
    \centering
    \newsavebox{\imgboxA}
    \sbox{\imgboxA}{%
      \includegraphics[width=0.9\textwidth]{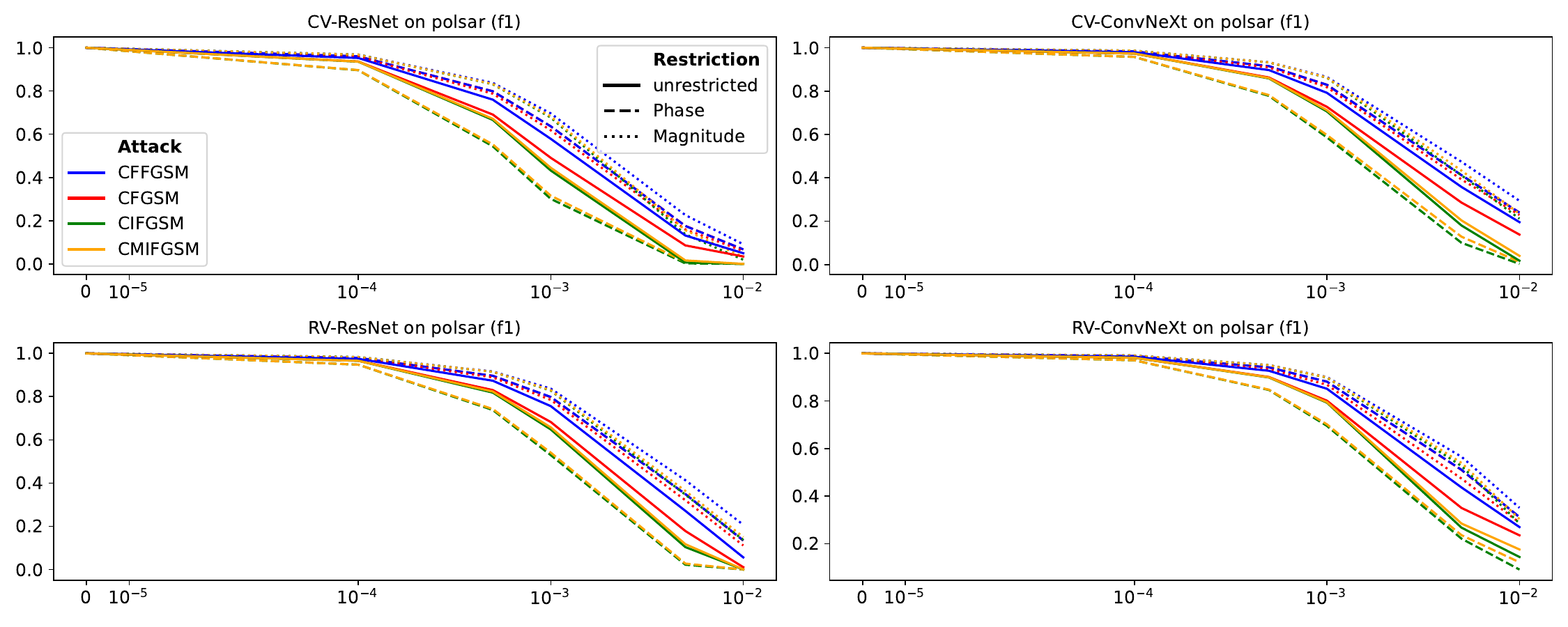}%
    }
    \newsavebox{\imgboxB}
    \sbox{\imgboxB}{%
      \includegraphics[width=0.9\textwidth]{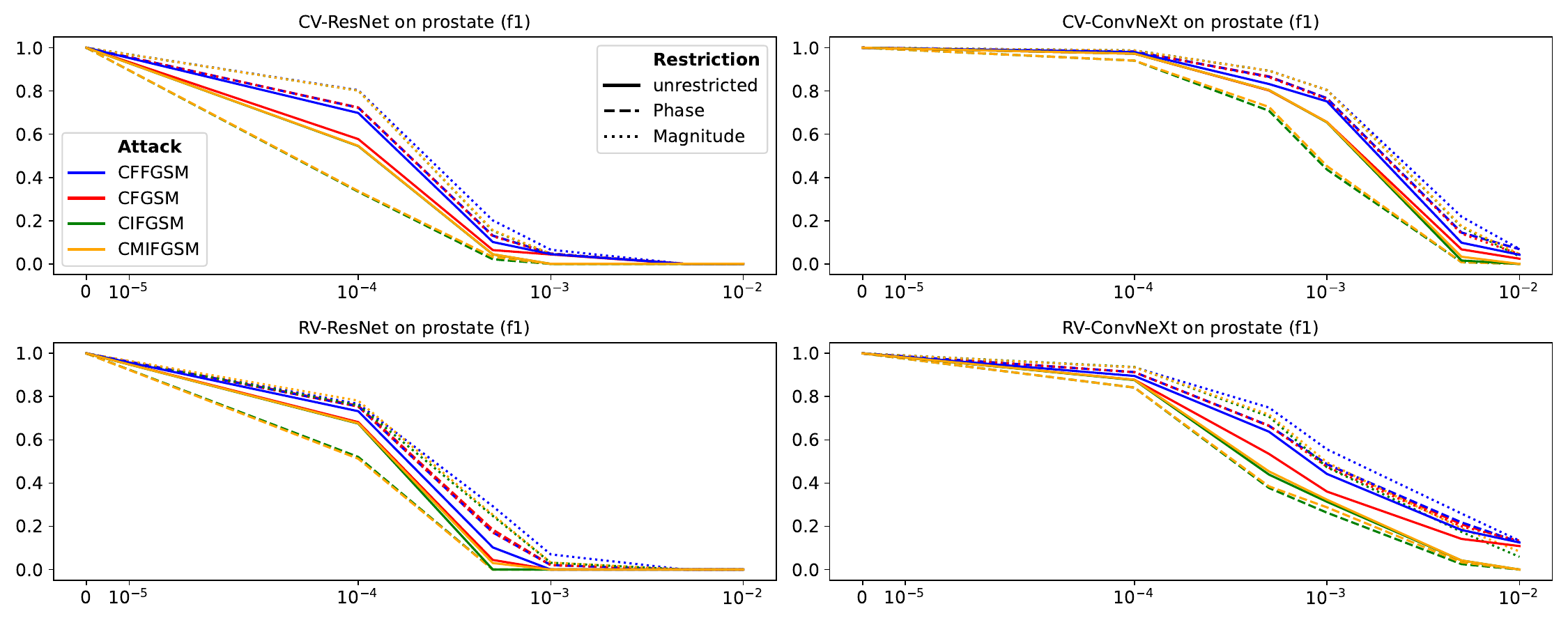}%
    }
    \newlength{\imgH}
    \setlength{\imgH}{\dimexpr\ht\imgboxA+\dp\imgboxA\relax}

    \setlength{\tabcolsep}{2pt}
    
    \begin{tabular}{@{}c c c@{}}
    \adjustbox{valign=c}{%
      \begin{minipage}[c][\imgH][c]{8pt}
        \centering
        \rotatebox[origin=c]{90}{PolSAR dataset}
      \end{minipage}%
    }
    &
    \adjustbox{valign=c}{%
      \begin{minipage}[c][\imgH][c]{8pt}
        \centering
        \rotatebox[origin=c]{90}{CVNN}\par\vspace{60pt}
        \rotatebox[origin=c]{90}{RVNN}
      \end{minipage}%
    }
    &
    \adjustbox{valign=c}{\usebox{\imgboxA}} \\
    \adjustbox{valign=c}{%
      \begin{minipage}[c][\imgH][c]{8pt}
        \centering
        \rotatebox[origin=c]{90}{Prostate dataset}
      \end{minipage}%
    }
    &
    \adjustbox{valign=c}{%
      \begin{minipage}[c][\imgH][c]{8pt}
        \centering
        \rotatebox[origin=c]{90}{CVNN}\par\vspace{60pt}
        \rotatebox[origin=c]{90}{RVNN}
      \end{minipage}%
    }
    &
    \adjustbox{valign=c}{\usebox{\imgboxB}}
    \end{tabular}
    
    \caption{Comparison of Phase Attacks and unrestricted attacks without adversarial training.
    Left: ResNet, Right: ConvNeXt.
    All plots show (normalized) performance against $\epsilon$-restriction.
    Linestyle: straight line: Regular Attack, dashed line: Phase Attack, dotted Line: Magnitude Attack.
    Colors differentiate attack algorithms as shown in legend.}
    \label{fig:phasevsbase}
\end{figure*}

The results on the models with additional adversarial training (Fig.~\ref{fig:cvvsreal}, third and fourth row) show a similar trend. It can be observed that the model performance stays more stable for attacks with a lower attack bound, but still decreases with increasing attack strength. 
Especially for the Prostate dataset, adversarial training has a greater effect on the CVNNs than on the RVNNs, leading to (relatively) improved robustness on both models. On the ResNet this leads to a more robust CVNN compared to the RVNN, while on the ConvNext the increased robustness is able to equalize the superior robustness of the RVNN without adversarial training.

\subsection{Comparing Phase, Magnitude and Regular Attacks} \label{subsec:basevsphase}

The results on both datasets show a similar pattern (Fig.~\ref{fig:phasevsbase}). The Phase Attacks, Magnitude Attacks and the unrestricted attacks are well suited to decrease the model performance, if the strength of the attack is sufficient. 

The Magnitude Attacks are generally less effective than the Phase Attacks and unrestricted attacks, following the latter's overall trends. Thus, we focus on discussing the differences between the Phase and Regular Attacks.

For both models and on both datasets the single-step Phase Attacks PFGSM and PFFGSM are less effective than their counterparts $\C$FGSM and $\C$FFGSM. Interestingly however, the multi step Phase Attacks PIFGSM and PMIFGSM are more effective than $\C$IFGSM and $\C$MIFGSM. On the PolSAR dataset the multi step Phase Attacks always accomplish the goal of completely fooling the model, i.e. reaching a score of $0$, on all tasks with a maximum perturbation of 0.01 or higher, while the non-restricted base attacks do never accomplish this with any of the given perturbation strengths. On the Prostate dataset the models performance decreases even more by both attacks, sometimes already reaching a score of $0$ in all cases with a perturbation strength of 0.005.

As adversarial training has no meaningful impact on the comparison of the three attacks, the results are omitted for the sake of brevity.
Notably, adversarial training is also effective against the Phase and Magnitude Attacks even though it was only done with the unrestricted attacks.
All observations made on the results without adversarial training still largely hold.

\subsection{Discussion}

The first set of experiments show that CVNNs are generally on par or more robust to adversarial attacks than RVNNs, regardless of adversarial training. There is only one case (ConvNeXt on prostate without adversarial training) in which the RVNN is more robust.
In all other scenarios, the performance of RVNNs decrease faster than the performance of the corresponding CVNNs, when applying the same adversarial attack.
Notably, generally even the weakest adversarial attack ($\C$FGSM) has a stronger effect on the RVNNs than the strongest attack ($\C$IFGSM) on the CVNNs. In two experiments however, the RVNN is more robust than the CVNN.

The second set of experiments show that both CVNNs as well as RVNNs are highly susceptible to changes in the phase alone.
Even when restricting the adversarial attacks to only alter the phase of the input, the adversarial attacks stay effective.
Interestingly, in all experiments the iterative Phase Attacks showed to be more effective than the non-restricted base attacks. This is surprising, since from a theoretical standpoint the Phase Attacks are strictly weaker than the non-restricted attacks. This suggests that the restriction offered by the Phase Attacks makes it easier for the iterative optimization algorithm to find a suitable minimum.
This observation is further reinforced by the results on the Magnitude Attacks against which the models are more robust than even the unrestricted attacks and therefore also the Phase Attacks.

\section{Conclusion}

In this work, we have shown evidence that CVNNs can be more robust to adversarial attacks than RVNNs in some cases.
To further investigate how adversarial attacks behave in the complex domain, we introduced Phase Attacks --- adversarial attacks, that are limited to attack the phase of the complex-valued input but leave the magnitude unaltered.
The Phase Attacks show a similar and in case of iterative optimizers even increased effectiveness when compared to regular attacks, leading to the conclusion that neural networks dealing with complex-valued data are highly susceptible to phase changes.
This is further reiterated by the fact, that the Magnitude Attack (which can not alter the phase at all) is less effective than the other two attack types.
This should be considered when dealing with complex-valued datasets, especially in safety critical environments such as MRI in clinical practice. 

As future work we would like to further investigate how CVNNs and RVNNs can be made more robust to phase changes specifically as well as test our hypothesis of susceptibility to phase perturbations on a wider scale. Additionally, the influence of changes in the phase on other definitions of robustness --- such as robustness to noise or distribution changes --- should be investigated.

\bibliographystyle{IEEEtran}
\bibliography{bibliography}
\end{document}